\documentclass[runningheads]{llncs}
\usepackage{graphicx}
\usepackage{amsmath,amssymb} 
\usepackage{color}
\usepackage{booktabs} 
\usepackage{floatrow}
\newfloatcommand{capbtabbox}{table}[][\FBwidth]
\usepackage{multirow}
\usepackage[font=small,labelfont=bf]{caption}
\usepackage{xcolor}
\usepackage{wrapfig}

\newcommand{\tinne}[1]{\textcolor{black}{#1}}
\newcommand{\tocheck}[1]{\textcolor{black}{#1}}

\begin{document}
\pagestyle{headings}
\mainmatter
\def\ACCV20SubNumber{106}  

\title{
Multiple Exemplars-based Hallucination \\for Face Super-resolution and Editing
} 
\titlerunning{\space}
\authorrunning{\space}

\author{Kaili Wang$\dagger$*, Jose Oramas*, Tinne Tuytelaars$\dagger$}
\institute{$\dagger$KU Leuven, ESAT-PSI,    *University of Antwerp, imec-IDLab}

\maketitle

\begin{abstract}
\tinne{Given a really low resolution input image of a face (say $16{\times} 16$ or $8{\times}8$ pixels), 
the goal of this paper is to reconstruct 
a high-resolution version thereof. This, by itself, is an ill-posed problem, as the high-frequency information
is missing in the low-resolution input and needs to be hallucinated, based on prior knowledge about the image content. 
Rather than relying on a generic face prior, in this paper we explore the use of a set of exemplars,
i.e. other high-resolution images of the same person. These guide the neural network as we condition the output on them. Multiple exemplars work better than a single one.}
To combine the information from multiple exemplars effectively, we introduce a pixel-wise weight generation module.
Besides standard face super-resolution, 
\tinne{our method allows to perform subtle face editing simply by replacing the exemplars with another set 
with different facial features.}
A 
user study is conducted and shows 
the super-resolved images can hardly be distinguished from real images on the CelebA dataset.
A qualitative comparison indicates our model outperforms methods proposed in the literature on the CelebA and WebFace datasets.






\end{abstract}

\section{Introduction}
\label{sec:intro}

\begin{figure}
\centering
\includegraphics[width=1\textwidth]{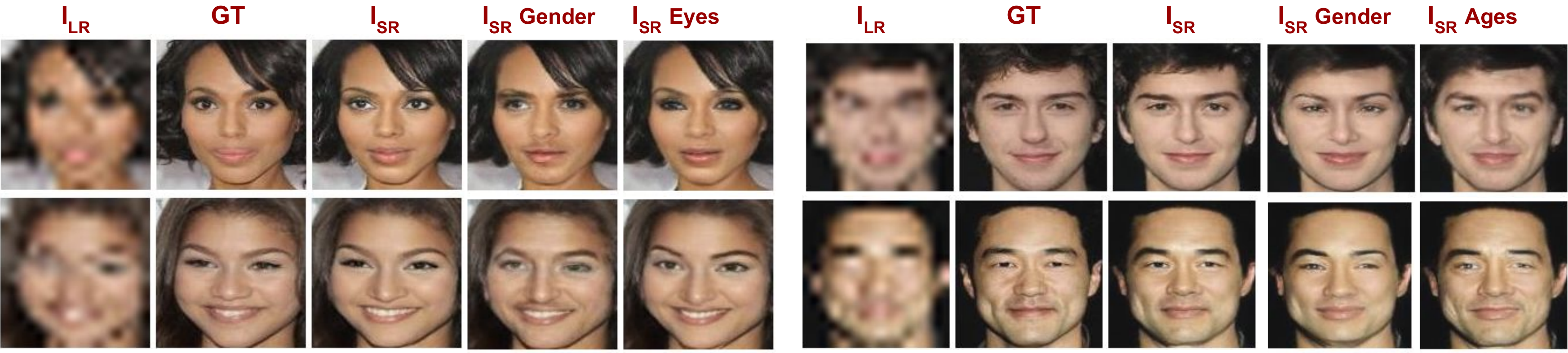}
\caption{Super-resolving images with different facial features by using different exemplars. From left to right: input LR image, ground truth HR image, SR image using exemplars of the same person and SR images using two sets of exemplars with different facial features. Please note, the ethnicity of the LR image does not change. For sake of space, please refer to the supplementary material for the exemplars used.}
\label{fig:teaser}
\end{figure}

Super resolution (SR) imaging consists of enhancing, or increasing, the resolution of an image, i.e., going from a coarse low-resolution (LR) input to one with high resolution (HR) depicting more details.
%
Especially challenging is the setting where there is a large scale factor
between the resolution of the input and that desired for the output. 
%
In that case, there is insufficient information in the LR input. This leads to the need to ``hallucinate"  what the detailed content of the HR output would look like. As can be seen in Fig.~\ref{fig:teaser} this makes the SR process an ill-defined problem with multiple valid HR solutions for a given LR input.

Using side-information is a promising avenue towards guiding the SR process and addressing
the above
issue. 
%
%
In particular, we consider using multiple HR reference images, or {\em exemplars} as we call them, as a form of side information.
\tinne{Typically, these are HR images of the same person depicted in the LR input.} They provide visual cues that are expected to occur in the HR output -- think e.g.~of the color or shape of the eyes of the person. 
By guiding the SR process these exemplars help to constrain the output and disambiguate the ill-defined characteristic of the problem. 
\tinne{The output becomes conditional on the provided exemplars and a different set of exemplars will lead to a different output.}

\tinne{In this work, we advocate the use of {\em multiple} exemplars, rather than just one. This gives more flexibility, as it allows the network to pick the visual features that best fit the LR input image,
in terms of facial expression, illumination or pose.}
%
Our method effectively exploits available information from the exemplars via the proposed Weighted Pixel Average (PWAve) module. This module learns how to select useful regions across the exemplars and produces superior results compared to simply averaging the representations~\cite{liu2019few}.

%
Rather than just superresolving the LR input,
we can also exploit the ambiguity of the problem to our advantage and use different sets of exemplars as a means to inject different visual features in the produced HR image (see again the examples in figure~\ref{fig:teaser}). This turns the model into a flexible tool for subtle image editing.
Since in this manuscript we focus on images depicting faces, we refer to the general SR process as "face/facial hallucination". Similarly, we refer as "face editing" to the task where new features are injected.
Investigating these two tasks constitutes the core of this paper.


\tinne{Taking a broader perspective, the method proposed here to condition a network 
on a set of exemplars, can be seen as a novel, lightweight and flexible scheme for model
personalization, that could have applications well beyond face hallucination or face editing. 
Indeed, instead of finetuning a model on user-specific data, which requires a large set
of labeled data and extra training, or adapting a model using domain adaptation techniques, which requires
access to the old training data as well as a large set of unlabeled target data and extra training, 
the scheme we propose allows to adapt a generic model to a specific user without the need
for any retraining. All that is needed is a small set of exemplars and a standard forward
pass over the network. This opens new perspectives in terms of on-the-edge applications,
where personalization is often a desirable property yet computational resources are limited.
}

Our contributions are three-fold: 
i) We propose to use multiple exemplars
to guide the model to super-resolve very low resolution ($16{\times}16$, $8{\times}8$) images.
ii) Our model does not require any domain-specific features, e.g. facial landmarks, in the training phase. This makes our model general to adapt to other datasets and tasks. Our user study indicates our results are hard to distinguish from real images. Our qualitative analysis suggests that our model outperforms the literature baselines.
iii) Besides achieving face super-resolution, our model can also address the face editing task. Unlike traditional conditional generative adversarial networks which can only generate images by modifying pre-defined discrete visual features, our model can dynamically generate HR face images with arbitrary facial features, benefiting from the usage of exemplars.
%
\begin{figure}
\centering
\vspace{-4mm}
\includegraphics[width=0.9\textwidth]{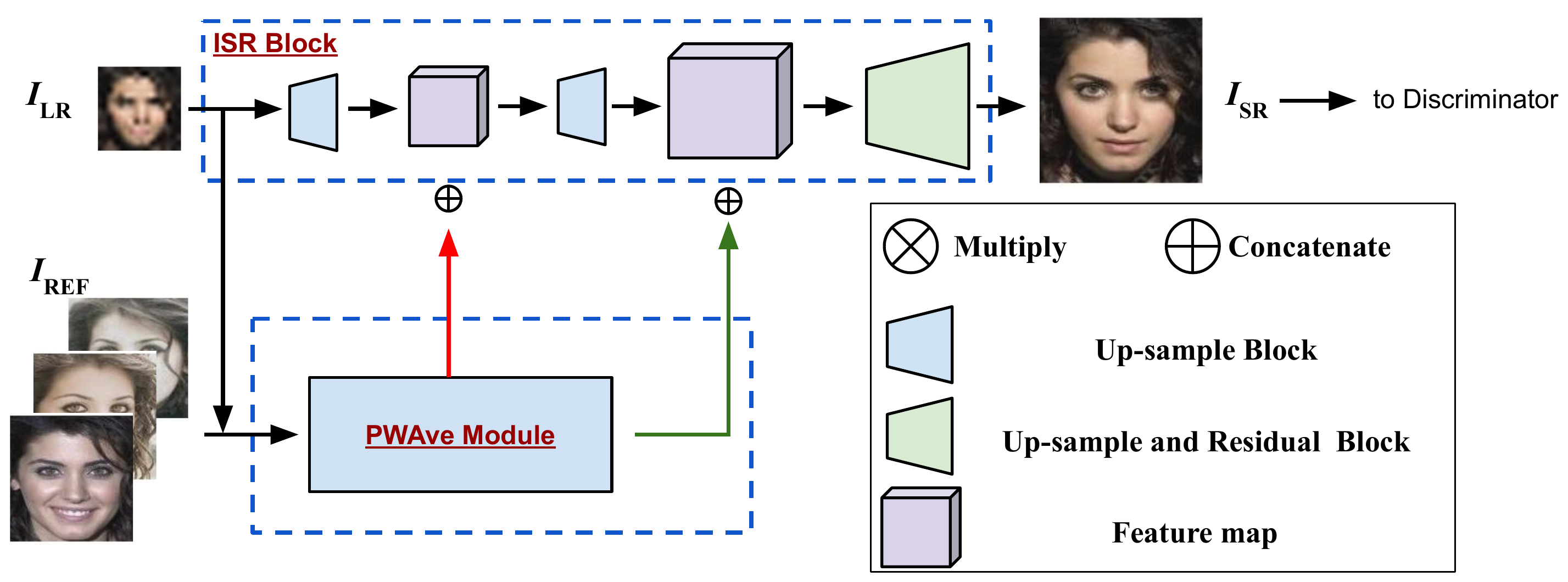}
\caption{The proposed model in our paper. For details of the PWAve module, please refer to Sec. \ref{sec:method} and figure~\ref{fig:pwaveModule}.}
\label{fig:proposedApproach}
\end{figure}
\vspace{-12mm}


\section{Related Work}
\label{sec:relatedWork}
\subsubsection{Face super-resolution}
Using deep neural networks to super-resolve LR images has been applied quite widely.
The methods can be gathered into two groups: those using facial priors like landmarks or heatmaps, and those not using any facial priors.
\cite{CT-FSRNet-2018,Super-FAN18,Yu_2018_ECCV,progressive-face-sr_19} 
belong to the first group.
Yet to use these facial priors, extra annotations and networks are required. 
In the other group, 
\cite{attribute_face_18_cvpr,Lu_2018_ECCV,FaceWarp_2018_ECCV,FaceWarp_2019_CVPR_Workshops,Xin_2019_residualAttri,zhang2018superidentity}
use a guidance-image or semantic labels (e.g. facial attributes or face identity) to help the model super-resolve the LR images. 
\cite{attribute_face_18_cvpr,Lu_2018_ECCV}  utilize pre-defined facial attribute labels. Besides super-resolving the LR image, as a conditional adversary model, it allows users to change the feature labels to control the generated SR image.
However, to achieve this, they require additional data with annotations for the visual attributes to be added/edited.
Furthermore, editing of visual features is also limited to the set of visual attributes pre-defined at design time, and fixed to the possible values that these features may take.
Moreover, when new features are desired to be integrated in the editing model, additional data and re-training are required.
In addition, \cite{attribute_face_18_cvpr} manually rotates the input LR image and uses a spatial transformer to cope with this change. Compared with \cite{attribute_face_18_cvpr,Lu_2018_ECCV}, our method can be regarded as conditioned on multiple exemplars, which gives us more flexibility on the subtle editing of facial features.
%
%
\cite{Xin_2019_residualAttri} does not use any guiding image, but uses the facial feature labels to train a channel-wise attention model which guides the model to recombine the basic features in the super resolution process. Compared with \cite{Xin_2019_residualAttri}, our method uses pixel-level weights generated from our PWAve module, which provides the user with a more clear visualization on the usage of exemplars. In addition, the way of using facial features in \cite{Xin_2019_residualAttri} does not allow users to control the generation.
\cite{zhang2018superidentity} considers the facial identity label, where they use a specific person identity loss. 
\cite{FaceWarp_2018_ECCV,FaceWarp_2019_CVPR_Workshops} use one guiding exemplar while we propose to use multiple guiding images. They warp the guiding exemplar (HR) via a flow field and use a specifically designed network to align with the input LR image.
\cite{FaceWarp_2018_ECCV} also uses facial landmarks in the training of the warping network.



\subsubsection{Conditional Image Generation}
\cite{ACGAN_17,cgan_2014,infogan_16,CVAE_17,attri2image_16} use labels as condition to guide neural networks to generate images. Therefore, the generated images are limited to the pre-defined classes.
\cite{liu2019few,esser2018variational,Lin_2018_CVPR,Huang_2018_ECCV} achieve image translation by using condition image(s). In the training process, they aim to disentangle the visual features from both input and condition images and re-combine the features from different images later. 
\cite{liu2019few} simply averages the features from condition images and uses adaptive instance normalization to insert them.
To achieve the modification of facial features, our method does not need to re-train the original SR model. Our SR model is originally trained to restore the LR image with the help of several exemplars that have similar visual features to the LR image. 
Besides, our PWAve module generates the pixel-wise weights among the multiple exemplars.
In addition, the input image in our method is a LR image while conditional image translation efforts~\cite{liu2019few,esser2018variational,Lin_2018_CVPR,Huang_2018_ECCV} receive the HR image as part of its input.

\section{Methodology}
\label{sec:method}
In this section, we describe the architecture of our model and the objective functions required for its optimization.
%
For each of the LR images to be super-resolved we assume the availability of a set of high-resolution exemplars. These exemplars contain a person with the same identity as the person depicted in the LR image. 
%
%
Specifically, we assume that the exemplars possess detailed facial features, e.g. eye shape, iris color, gender etc. that 
are expected to appear in the SR image.
These exemplars are used to provide the network with these facial features that are invisible in the LR image.

\subsection{Architecture}
The generator model consists of two parts: the image super resolution block (ISR) and the Weighted Pixel Average (PWAve) module (see Figure~\ref{fig:proposedApproach}).
The ISR block contains $M_1$ up-sample blocks and $M_2$ residual blocks .
%
%
The ISR block takes the LR image ($I_{LR}$) as input and integrates at different parts of its feed-forward pass combined feature maps from the exemplars as generated by the PWAve module. 


Besides face super-resolution, the combination of feature maps from the input LR image and the exemplars allows the model to dynamically generate HR images with similar facial features to those present in the exemplars.
%
\begin{figure}
\centering
\includegraphics[width=0.7\textwidth]{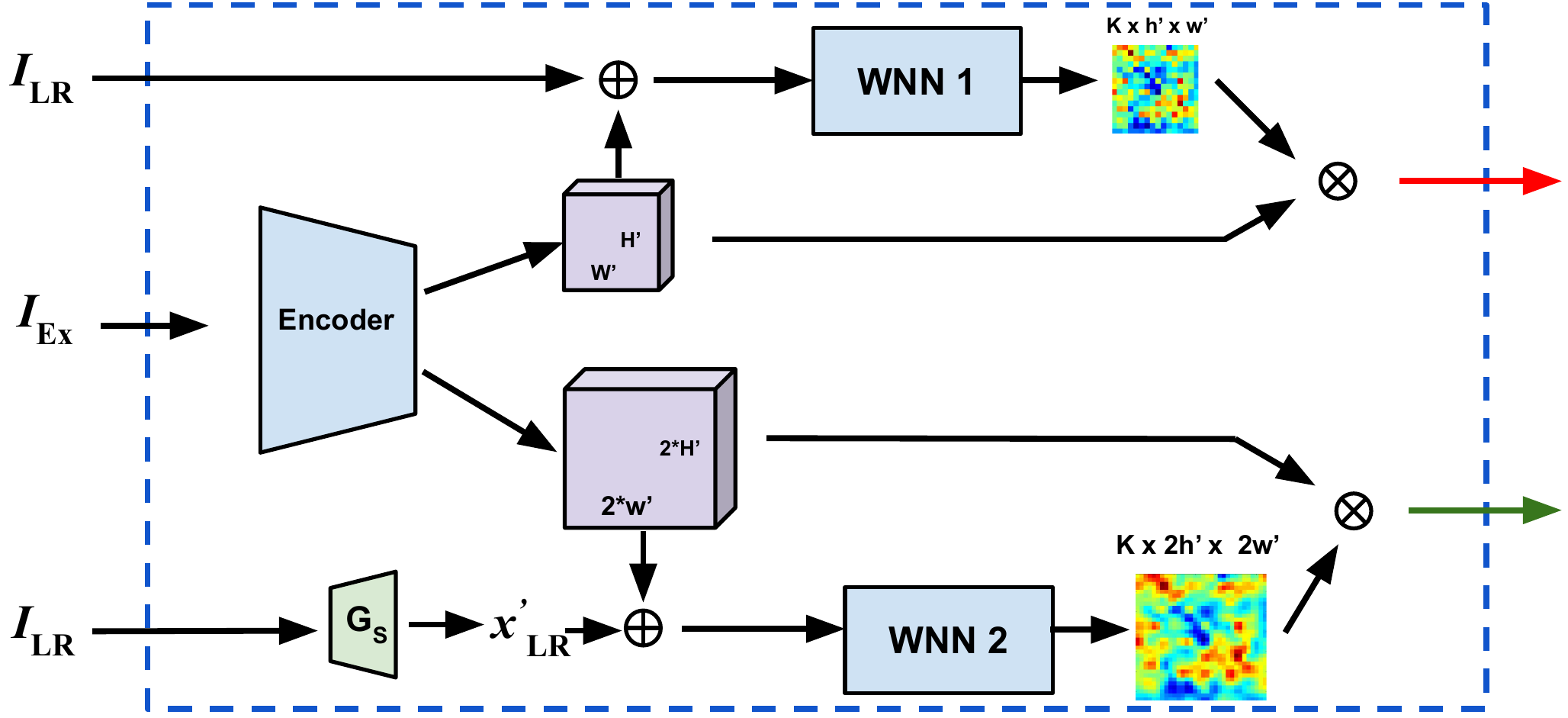}
\caption{The proposed PWAve module in our paper. To concatenate with two times higher resolution feature map, we train a small generator to get $x^{'}_{LR}$. For the whole model, please see Fig. \ref{fig:proposedApproach}.}
\label{fig:pwaveModule}
\end{figure}
%
\subsubsection{Weighted Pixel Average (PWAve) module}
The goal of this module is to learn how to perform a good combination of the set of intermediate feature maps computed from the exemplars. The idea is to provide the freedom to the model to learn a different, and perhaps more suitable, combination method rather than the commonly used average method~\cite{liu2019few}. 
%
%
Our intuition is that the exemplars could depict different features, e.g. facial expression, angle, makeup, etc., that are in the LR image and are expected to appear in the SR image.
Therefore, it is meaningful to consider the exemplars when the model generates the HR image so that the model can take into account proper regions of the exemplars. 

The PWAve module consists of an encoder $E_{Ex}$ and $M$ pixel weight generation networks (WNN). 
Here, we set $M {=} 2$.
$E_{Ex}$ encodes the exemplars 
$I_{Ex} \in \mathbb{R}^{N\times K\times 3 \times h \times w}$, with width $w$ and height $h$,
into feature maps 
$f_{Ex} \in \mathbb{R}^{N\times K\times C \times h' \times w'}$ 
of different scales, $N$ and $K$ are the batch size and the number of exemplars, respectively. We use two scales: the same scale with and two times higher than $I_{LR}$. WNN takes $f_{Ex}$ and $I_{LR}$ of corresponding size and generates the pixel-wise weight matrix $W$ (
$W \in \mathbb{R}^{N\times K \times1 \times h' \times w'}$ ) across the exemplars.
Afterwards, 
$W$ is $L_1$-normalized along the second dimension. In order to concatenate with the two times higher resolution feature maps and provide a more precise guidance in WNN-2, we use a small generator $G_S$ to generate $I_{LR}^{'}$ rather than just up-sample $I_{LR}$.
Finally, the combined exemplar feature maps $f_{Ex}^{c}$ are calculated by applying dot-product between $f_{Ex}$ and $W$, followed by the summation along the second channel, i.e. $f_{Ex}^{c} \in \mathbb{R}^{N\times 1\times C \times h' \times w'}$. 
Different from \cite{liu_2017_qan}, we generate pixel-wise weights while taking into account the LR image itself. Taking into account $I_{LR}$ (and $I_{LR}^{'}$) in the combination process is a more direct way to identify the important region on the exemplars compared with simply relying on loss penalty.

\subsubsection{Discriminator (Critic)}
The architecture of the discriminator (critic) is similar to StyleGAN\cite{stylegan19}, where it takes either super-resolved images or original images as input and tries to match the distribution between the super-resolved image and real ground-truth HR images.

\subsection{Objective Functions}
We formulate this task as a supervised learning problem given the simplicity to obtain a low/high resolution image pair from a single high-resolution image.
\subsubsection{Content Loss ($L_{c}$)} We apply a $L1$ loss on the ground truth HR image $I_{HR}$ and the super-resolved image $I_{SR}$. 
\begin{align}
    \mathcal{L}_{\text{c}} = \| I_{SR} - I_{HR} \|_1
    \label{eq:CC_loss}
\end{align}
We also calculate the content loss for $G_S$, 
\begin{align}
    \mathcal{L}_{\text{c}}^{\text{s}} = \| I_{SR}^{s} - I_{HR}^{d} \|_1
    \label{eq:CC_loss}
\end{align}
where $I_{HR}^{d}$ and $I_{SR}^{s}$ are the real image that is downsampled to match the resolution of output of $G_S$ and the output of $G_S$, respectively.
\subsubsection{Perceptual Loss ($L_p$)} This loss aims at preserving the face appearance and identity information of the depicted persons. 
We use the perceptual similarity model $\Phi_p$ \cite{zhang2018perceptual}, which is trained on an external dataset, and the last layer before the classification layer of the face emotion model $\Phi_{ID}$ \cite{Albanie18a} to calculate the perceptual loss.

\begin{align}
    \mathcal{L}_{\text{p}} = \| \Phi_{p}(I_{SR})-\Phi_{p}(I_{HR}) \|_2^2 + \| \Phi_{ID}(I_{SR})-\Phi_{ID}(I_{HR})\|_2^2
    \label{eq:CC_loss}
\end{align}
where $I_{HR}$ and $I_{SR}$ are the real HR image and corresponding super-resolved HR image, respectively.
Similarly, we apply the perceptual loss on $I_{LR}^{'}$ and $I_{SR}^{d}$. 

\begin{align}
    \mathcal{L}_{\text{p}}^{s} = \| \Phi_{p}(I_{SR}^{s})-\Phi_{p}(I_{HR}^{d}) \|_2^2 + \| \Phi_{ID}(I_{SR}^{s})-\Phi_{ID}(I_{HR}^{d})\|_2^2
    \label{eq:CC_loss}
\end{align}

\subsubsection{Adversary Loss ($L_{adv}$)}
In order to generate a more realistic and sharper image, we use the adversary loss to match data's distribution. Furthermore, here we use Wasserstein GAN with gradient penalty (WGAN-GP) \cite{NIPS2017_wgangp} in order to have a more stable training process. 
The critic loss, used to update the discriminator, is defined as follows:
\begin{align}
    \mathcal{L}_{\text{critic}} = D(I_{SR}) + D(I_{HR}) - \lambda_{gp}\mathcal{L}_{gp}
    \label{eq:CC_loss}
\end{align}
where $D(.)$ is the discriminator (critic) network and $\lambda_{gp}$ is the coefficient parameter for gradient penalty. 
The adversary loss for the generator part is $-D(I_{SR})$. 

\subsubsection{Total loss ($L_{total}$)}
With the previous terms in place, the total loss is defined as:
\begin{align}
    \mathcal{L}_{\text{total}} = \big[\mathcal{L}_{c} + \lambda_1\mathcal{L}_{p} + \lambda_2\mathcal{L}_{adv} \big] + \big[\mathcal{L}_c^s + \lambda_3\mathcal{L}_p^s\big]
    \label{eq:CC_loss}
\end{align}
where $\lambda_{1}$, $\lambda_{2}$ and $\lambda_{3}$ are the trade-off coefficients of the model.
The losses in the first bracket will update the whole model except for $G_S$, where $G_S$ is updated by the losses in the second bracket.

\section{Evaluation}
\label{sec:experiment}
In this section, we start with the introduction of the dataset used in our experiments. This is followed by qualitative and quantitative results obtained by our method. We also conduct a user study on the quality of our super-resolved images. Then, we perform a qualitative comparison of our method w.r.t. methods from the literature. In addition, we conduct an ablation study to show the effectiveness of the number of exemplars and the proposed PWAve module. 
Finally, we show our method is capable of dynamically introducing features on the generated images via the exemplars.

\subsubsection{Dataset}
We use two datasets depicting human faces: CelebA\cite{celeba} and WebFace\cite{webface}.
CelebA has 202,599 images with 10,177 identities, each identity has 20 images on average. We follow \cite{attribute_face_18_cvpr} and crop the images to the size of 128${\times}$128 and drop the identities which have less than 5 images.
We downsample the original 128${\times}$128 image to the size of 16${\times}$16 and 8${\times}$8. We also follow the training/testing splits in the original CelebA dataset. There is no identity overlapping between these splits.
Images from the WebFace dataset have a higher 256${\times}$256 resolution.
There are 10,575 identities and each of them has several images, ranging from 2 to 804. For each identity, we follow \cite{FaceWarp_2018_ECCV} and select the top 10 best quality images. 
We drop the identities which have less than 10 images.
The images of the first 9121 identities are used for training and the rest of them for testing.
Unlike \cite{FaceWarp_2018_ECCV}, for both datasets, we select the LR/HR images and the corresponding exemplars at random. That is, there is no constraint on the facial angle and expression.

\subsubsection{Implementation details}
We implement our model in PyTorch\cite{pytorch2017}.
Height and width of $I_{LR}$ is $h_{LR}$ and $h_{LR}$.
For the CelebA dataset with the scaling factor of 8/16, the setting is $M_1 {=} 3/4$, $M_2 {=} 1/1$, $h_{LR} {=} w_{LR} {=} 16/8$, $N{=}8$.
For the WebFace dataset with the scaling factor of 8/16, the setting is $M_1 {=} 3/4$, $M_2 {=} 1/1$, $h_{LR} {=} w_{LR} {=} 32/16$, $N{=}4$.
We use Adam \cite{kingma2014adam} optimizer with $\beta_1 {=} 0$ and $\beta_2 {=} 0.99$. 
The initial learning rate is set to 0.003 for the whole model except for the two WNNs, whose initial learning rate is 0.0001. 
Please refer to the supplementary material for more details.
\subsection{Quality and quantity results}
\label{sec:qualityQuantityOurResults}

\begin{figure}
\centering
\vspace{-4mm}
\includegraphics[width=1\textwidth]{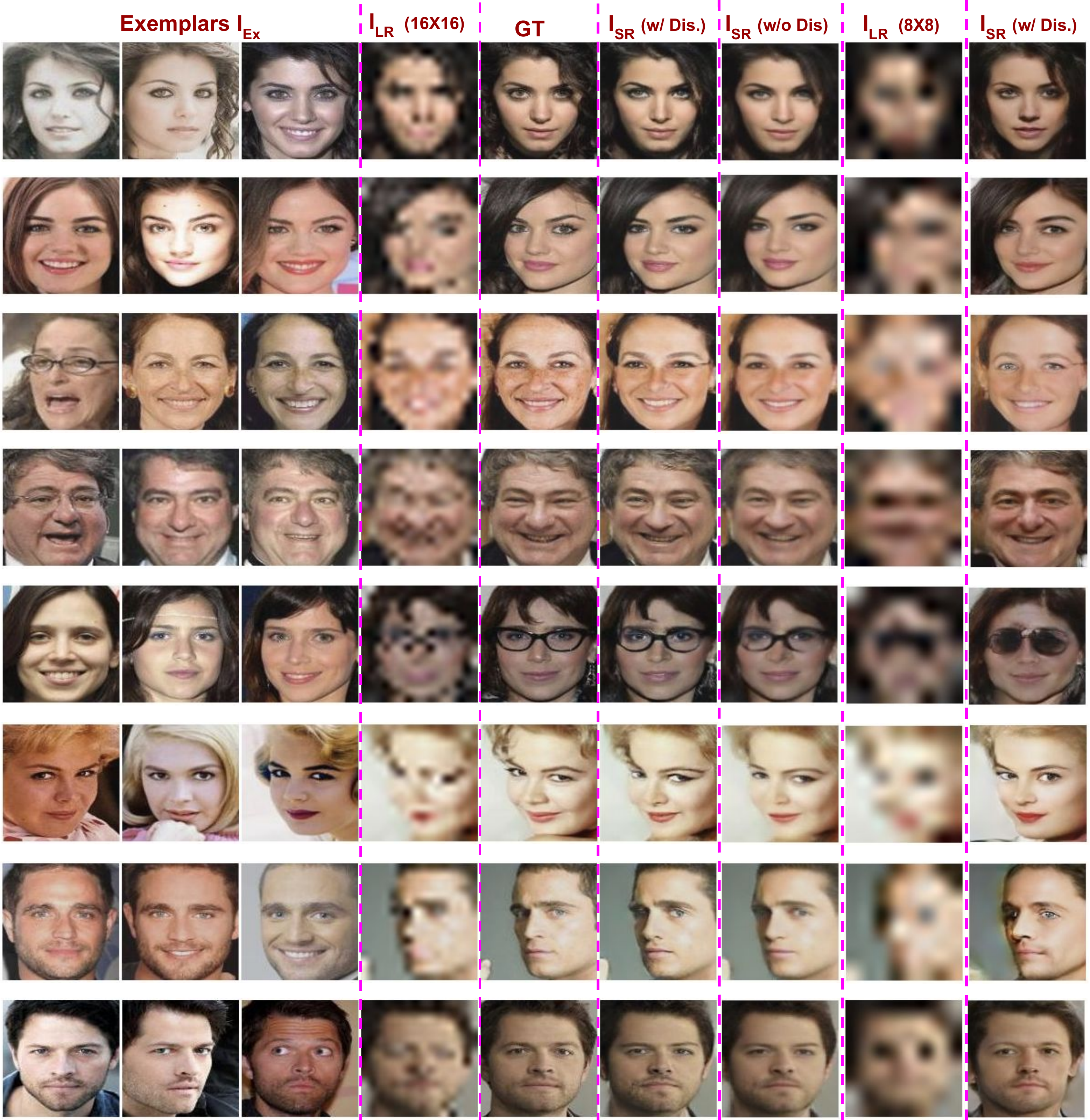}
\caption{Qualitative results from the CelebA dataset. We show two scaling factors: $\times$8 and $\times$16. The resolution of HR images is $128{\times} 128$. All the images are from testing set. }
\label{fig:celebAQuality}
\end{figure}

\begin{figure}
\centering
\includegraphics[width=1\textwidth]{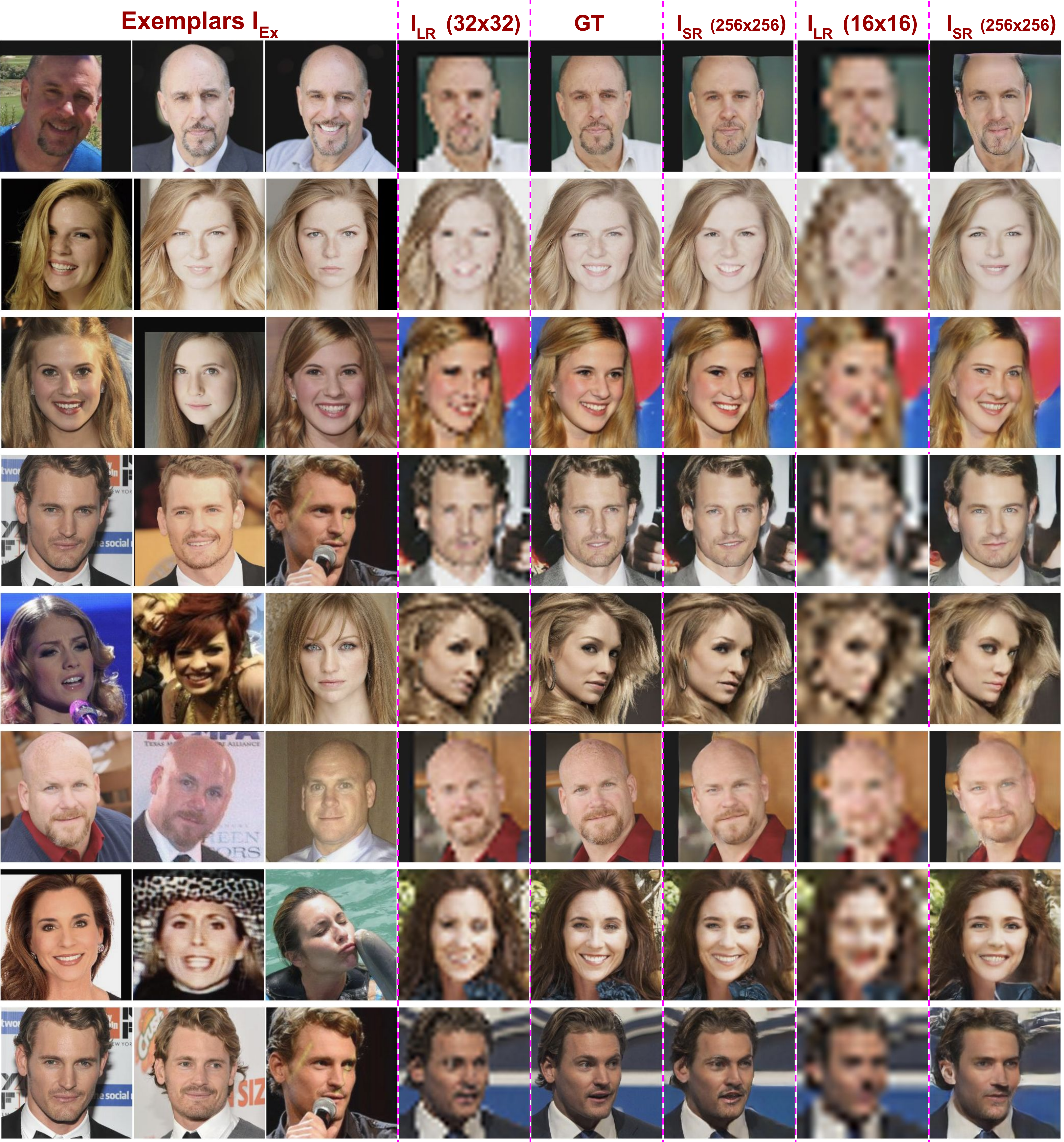}
\caption{Qualitative results on the WebFace dataset. We show two scaling factors: $\times$8 and $\times$16. The resolution of HR images is $256\times 256$ All the images are from testing set.}
\label{fig:webfaceQuality}
\end{figure}

Fig. \ref{fig:celebAQuality} and Fig. \ref{fig:webfaceQuality} shows qualitative results from the celebA and WebFace dataset. 
It is clear that our model can recover most of the details, such as wrinkles, iris color and stubble. In addition, our model is robust to changes in face angle, i.e. not only the front view but side view images can be super-revolved as well. Even when a scaling factor of 16 is used, it is interesting to see that the generated images (the last column of Fig.~\ref{fig:celebAQuality}) can also keep the original identity and details for most of the cases. 
\tocheck{For the fifth example in Fig. \ref{fig:celebAQuality}, the $8\times 8$ LR image has a black region around the eyes due to the downsampling of 16 times, which suggests the model to generate the sunglasses.}

\tocheck{For the WebFace dataset with the scaling factor of 16, the task is harder since the original size of the image is larger ($256{\times}256$), they are not aligned and have much more non-facial regions. However, the super-resolved images are still reasonable good based on the $16\times 16$ input and the corresponding exemplars.}
Table \ref{tab:celebAScore} shows Structural Similarity Index (SSIM) and Peak Signal-to-Noise Ratio (PSNR) scores on the testing set.

\vspace{+4mm}

\vspace{-4mm}
\begin{table*}
\setlength{\tabcolsep}{4.7pt} 
\centering
\caption{SSIM and PSNR scores on CelebA and WebFace dataset.}
\scalebox{0.9}{%
\begin{tabular}{l c c }
\toprule 
{Method} & SSIM & PSNR(dB) \\
& (CelebA/WebFace) & (CelebA/WebFace)\\
\midrule[0.6pt]
Bicubic ($\times 8$) &0.61/0.68 & 20.72/22.04 \\
Ours($\times 8$, w/o Discriminator) & 0.74/0.78& 23.45/24.92 \\
Ours($\times 8$, w/ Discriminator) &0.72/0.76 & 23.18/24.38
\\
\midrule[0.6pt]
Bicubic ($\times 16$) & 0.48/0.57&17.56/19.04 \\
Ours ($\times 16$ w/o Discriminator) & 0.62/0.67 &19.82/20.55 \\
Ours ($\times 16$ w/ Discriminator) &0.59/0.61 &19.13/19.42 \\
\bottomrule[1pt]
\end{tabular}
}
\label{tab:celebAScore}
\vspace{-2mm}
\end{table*}
\vspace{-4mm}

\begin{figure}
\centering
\vspace{-4mm}
\includegraphics[width=1\textwidth]{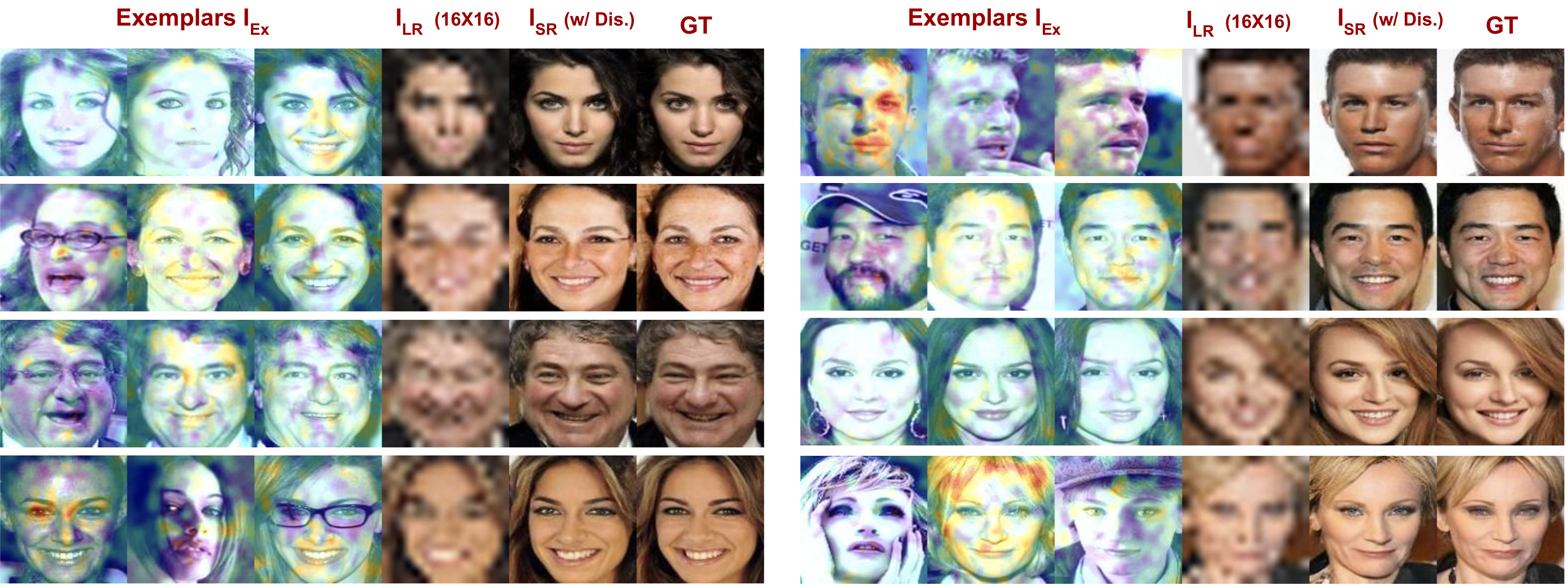}
\caption{The visualization of $W$ (Here $W \in R^{1\times 3\times 1\times16\times 16})$ generated by the PWAve module. Please note $W$ is normalized along the second channel. The heatmap is in jet color space, the warmer the color, the higher the weight.}
\label{fig:celebAHeatmap}
\end{figure}

%
To assess how the PWAve module helps the network in the generation process,
we visualize the weight matrix $W$, generated by the PWAve module, using the jet color scale and overlaying it on top of each of the exemplars. Fig. \ref{fig:celebAHeatmap} shows some examples.
The generated heatmaps clearly show that the PWAve module does learn how to select different parts from the exemplars rather than doing a random selection or a uniform average across images.
For example, the second row in the left part of this figure, the eyes and mouth parts of the first exemplar image have very low weight. This is reasonable since the face of the target image has a more frontal angle and does not have glasses.
If both the exemplar and target images have similar face angle, expression, etc., the PWAve module will generate weights that resemble an average operation. This can be seen in the third row on the right part.
\vspace{-8mm}

\subsection{User study}
\label{sec:userStudy}
\begin{wraptable}{r}{5cm}
\vspace{-8mm}
\caption{Results from our user study on the celebA dataset.}
 \resizebox{0.9\columnwidth}{!}{%
\begin{tabular}{l|c}
\toprule 
{Error rate ($\%$)} & 37.70 $\pm$12.53  \\
TPR ($\%$)          &  59.79$\pm$15.35  \\  
TNR ($\%$)          &  66.80$\pm$17.80  \\   
FPR ($\%$)          &  33.34$\pm$17.85  \\   
FNP ($\%$)          &  37.29$\pm$19.30 \\ 
\bottomrule
%
\end{tabular}
}
\label{tab:userStudy}
\end{wraptable}

We conduct an user study in order to quantitatively assess the quality of our super-resolved images as perceived by humans. 
Our survey was joined by 51 participants from around the world. From these participants, 35 have experience in computer science or informatics. We refer to this group as "CS". The rest of them do not have such experience, we refer to them as "Non-CS".

The survey was composed by two parts. The first part aims to check whether the participants could distinguish between the super-resolved images from real GT (HR) images. For each question, we put one HR image (128${\times}$128) next to the upsampled version of the corresponding LR image (16${\times}$16). Then we ask users to judge whether the shown HR image is the real, i.e. original, one.
The HR image is randomly sampled from the ground truth and the images generated by our method.
All the generated images shown on the survey are randomly sampled from our testing set.
In total, there are 200 questions, half of them with a ground truth HR image. For each survey we randomly sample 25 questions.
Before starting the survey, we also "train" the users by showing 12 examples with label (Real or Fake) in order to make them familiar with the task at hand.

The second part is more subjective, it consists on asking the participants to judge which type of the super-resolved images they consider more realistic: the sharp one (with Discriminator) or the smooth one (without Discriminator).

Table~\ref{tab:userStudy} shows the results of the first part of the study. The error rate is close to the random guess (50$\%$). The FPR and FNR are quite balanced.
In terms of the two user groups, CS group gets (34.57$\pm$11.22)$\%$ error rate while the Non-CS group achieves (43.62$\pm$13.51)$\%$, respectively.
As for the second part, although the SSIM and PSNR scores are lower (Table~\ref{tab:celebAScore}), around 60$\%$ of the participants think the sharp images look more realistic than the smooth ones.
Therefore, we will focus on the sharper qualitative results in the next sections.












\subsection{Qualitative comparison with respect to the state-of-the-art}
\label{sec:qualityCompare}

\subsubsection{Why there is no quantitative comparison?}
We do not provide a quantitative comparison w.r.t. existing methods for the following reasons: 
i) The SSIM and PSNR scores do not provide a definitive answer: blurry images may have a higher SSIM and PSNR score \cite{johnson2016perceptual}\cite{psnr_defect} since the optimal solution to minimize reconstruction error in image space is averaging all possible solutions~\cite{FaceWarp_2019_CVPR_Workshops}\cite{dosovitskiy2016generating}\cite{Ledig2017}.  
This observation was further ratified in our previous experiment.
ii) Existing methods are implemented in different frameworks and, currently, there is no unified benchmark for this task. Re-implementation may raise doubt on the quality of the code, and the used data splits and pre-processing.
Therefore, we only provide the conducted user-study (Sec.~\ref{sec:userStudy}) and a qualitative comparison. 

\begin{figure}
\centering
\includegraphics[width=1\textwidth]{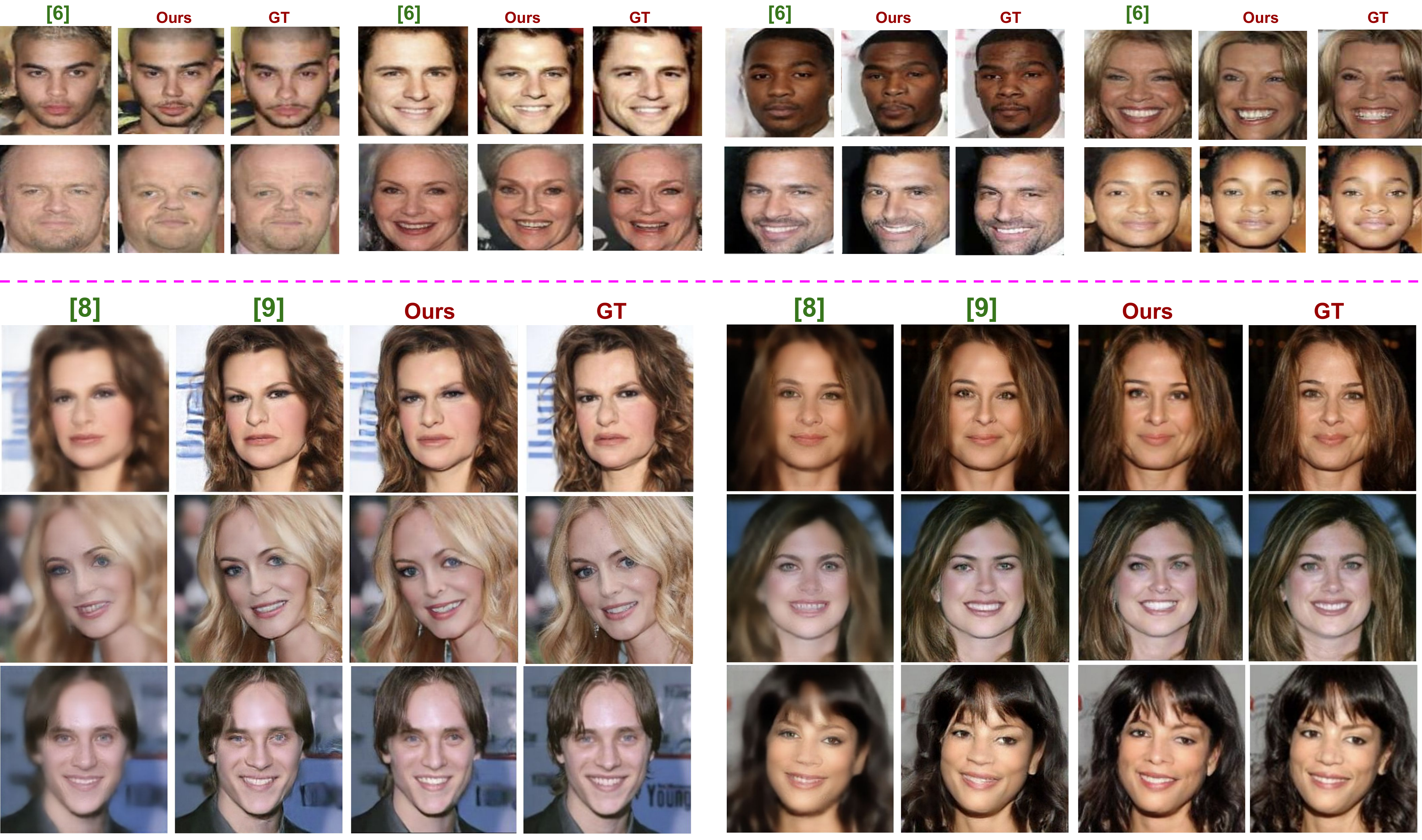}
\caption{Top: Qualitative comparison w.r.t. state-of-the-art methods on the celebA dataset. The resolution of $I_{LR}$ is 16${\times}$16 and the HR image is 128$\times$128. The first column shows the results from \cite{attribute_face_18_cvpr}. 
Bottom: Qualitatively comparison w.r.t. state-of-the-art baselines on WebFace dataset. The $I_{LR}$ has the resolution of 32 $\times$ 32 and HR image is 256 $\times$ 256. The first and second column are results from \cite{FaceWarp_2018_ECCV} and \cite{FaceWarp_2019_CVPR_Workshops}.}
\label{fig:webfaceComparison}
\end{figure}

In this experiment, the presented qualitative results (images) are taken directly from the corresponding papers. 
These selected images will constitute the point of comparison.
%
Fig.~\ref{fig:webfaceComparison} shows a qualitative comparison on celebA dataset w.r.t. \cite{attribute_face_18_cvpr} and on the WebFace dataset w.r.t. \cite{FaceWarp_2018_ECCV} and \cite{FaceWarp_2019_CVPR_Workshops}.
Generally, our model can generate more detailed results and preserve the person's original identity. 
\cite{attribute_face_18_cvpr} uses a manually rotated LR image as input while a few spatial transform networks are utilized to compensate this manual rotation. 
The results from \cite{attribute_face_18_cvpr} change the identity and the face shape for some cases. Moreover, some details are missing, e.g. the moustache is lost during the super resolution process in the first image in the top part of
Fig.~\ref{fig:webfaceComparison}.
The images generated by \cite{FaceWarp_2018_ECCV} have blurry hair as well as a blurry background. \cite{FaceWarp_2019_CVPR_Workshops} overcomes this problem but also loses some details, e.g. the eyes' shape (i.e. the third row in the bottom left of Fig.~\ref{fig:webfaceComparison}) and the iris color (i.e. the second row in the bottom right of Fig.~\ref{fig:webfaceComparison}). 

\subsection{Ablation study}
\label{sec:ablation}
\begin{figure}
\centering
\vspace{-2mm}
\includegraphics[width=0.98\textwidth]{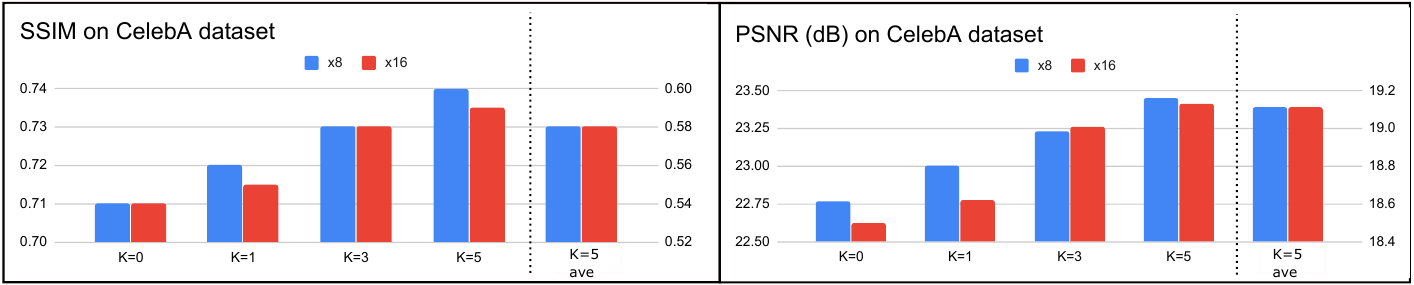}
\caption{Ablation study on the number of exemplars and feature map fusion method. The blue bar and the left y-axis are for the experiment with the scaling factor of 8 while the red bar and the right y-axis are for 16. Higher values are better.}
\label{fig:ablation}
\end{figure}

Our ablation study aims at assessing the effect of i) the number of considered exemplars, and ii) the PWAve module. Towards this goal, we conduct two experiments: gradually increase the number of exemplars $K$, from 0 (no exemplar) to 5, and replace the PWAve module by simply averaging the feature maps of exemplars. 
Fig~\ref{fig:ablation} shows the SSIM and PSNR scores under each setting. The trend is clear - more exemplars bring more benefits. The model with scaling factor of 16 can gain more, it is because the $I_{LR}$ contains less information and the model can gain extra information from the exemplars.

\begin{figure}
\centering
\includegraphics[width=1\textwidth]{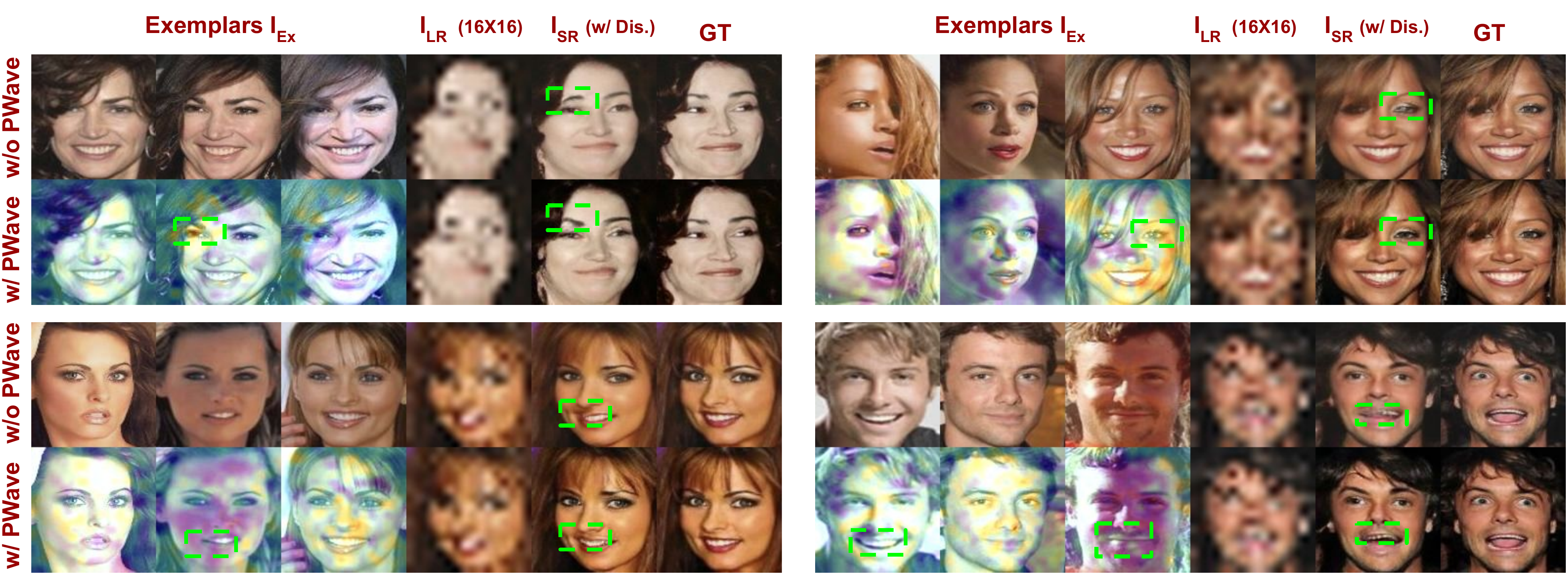}
\caption{Examples of ablation study on the PWAve module. For each set, the first row is without PWAve module (standard average). We show the heatmap of $W$ generated by PWAve on each the second row. Please give your attention to the region within the dashed green box highlighting keys aspects of the combination process.}
\label{fig:pwaveCompare}
\end{figure}

Regarding the PWAve module, the scores are also better than the averaging method. To give a  more straightforward comparison on this module, we show some examples in Fig.~\ref{fig:pwaveCompare}.
These examples show that the combination of the feature maps will influence the details in the generated image, such as mouth, eyebrows, etc.
With the help of the weights generated by the PWAve module, the model can take into account useful regions from the exemplars and produce a HR image closer to the ground truth.

\subsection{Facial features editing via exemplars}
\label{sec:faceEditing}
Based on the assumption that for most cases, the exemplars possess facial features such as eye shape, iris color, gender etc. that are in the LR image and are expected to appear in the SR image.
we have shown that the exemplars do provide useful information to help the model super-resolve the LR image.
This observation raises the question, what if we use exemplars with different facial features? Will the super-resolved images still contain the original features from the LR image or adopt the new features from the exemplars?
In the CelebA dataset, we still find some cases where the facial features within the same identity are different.
We run two experiments considering exemplars with different and the same identity. 
Please note, in this experiment we just replace the exemplars in the testing phase, no re-training is required.

\begin{figure}[t]
\centering
\includegraphics[width=1\textwidth]{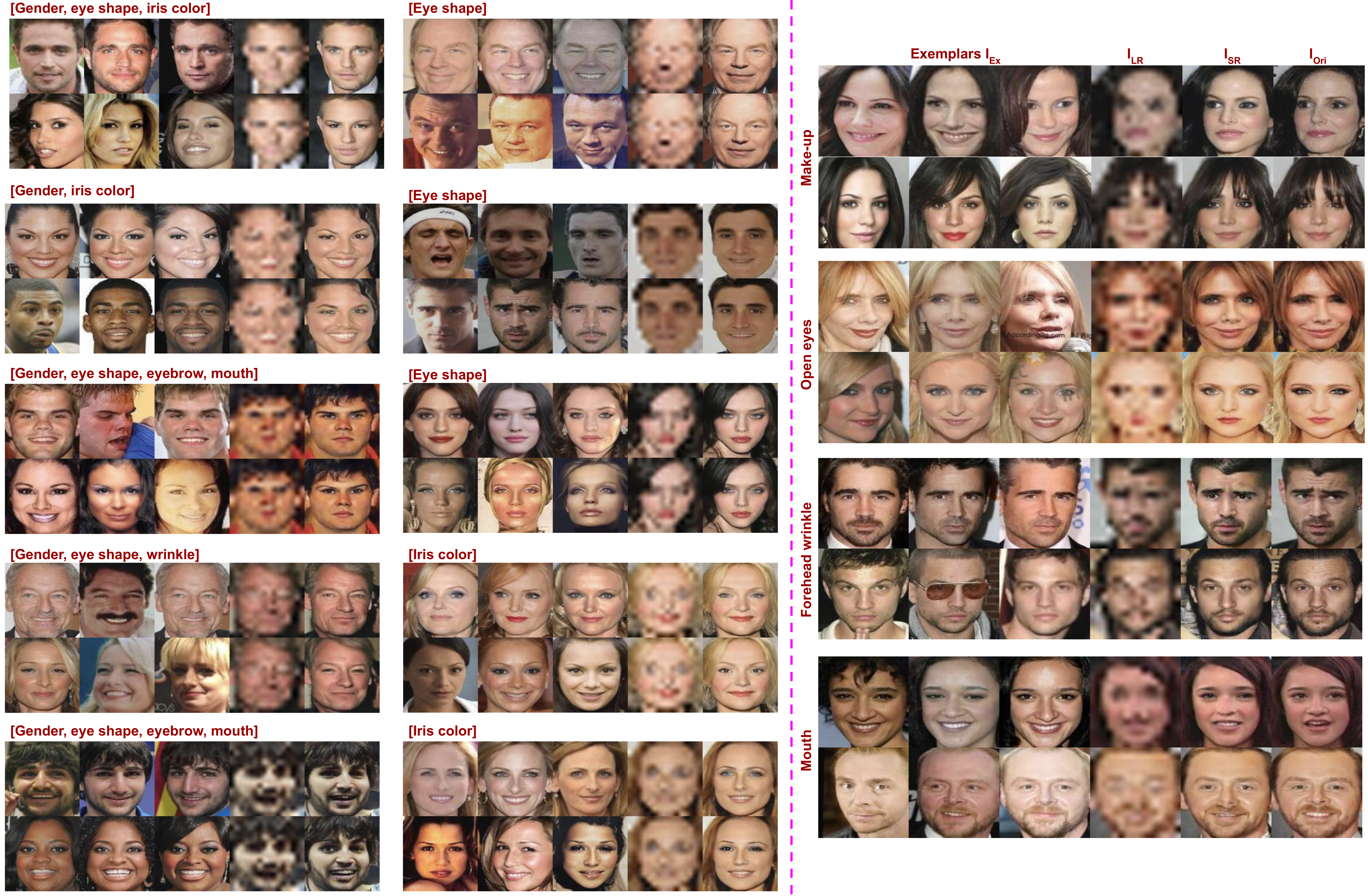}
\caption{Examples of editing/modifying facial features via exemplars. The left part shows example using $I_{Ex}$ with different identity. For each set, the first three images are the $I_{Ex}$, followed by the $I_{LR}$ and $I_{SR}$. The edited facial features are displayed on the top of each set. On the right part, $I_{Ex}$ has the same identity but different facial features. The edited facial feature is displayed on the left vertically.}
\label{fig:changeFaceAttri}
\end{figure}

Fig.~\ref{fig:changeFaceAttri} clearly answers the question above.
For both experiments, the super-resolved images adopt the features from exemplars, such as gender, makeup, eyes, mouth etc. 
More specifically, 
if the gender of exemplars is different (left part of Fig.~\ref{fig:changeFaceAttri}), the model will change the gender as well as other facial features.
We believe changing gender is more difficult since it is a high-level characteristic which is related to other low-level attributes such as eyes, eyebrows, mouth etc.
If the gender of the exemplars is the same (right part of Fig.~\ref{fig:changeFaceAttri}), then the identity of the original HR image will be maintained but changes in corresponding facial features occur.
Compared with \cite{attribute_face_18_cvpr}, 
benefiting from using the exemplars,
our method can generate face HR image with arbitrary facial features rather than only the pre-defined ones.  
The experiment shows that our model is capable of dynamically introducing features on the generated images via the exemplars. This suggests that our model can be also applied to subtle image editing tasks without the need of re-retraining or additional modifications on the original model. 


\section{Conclusion}
In this paper, we propose to use multiple exemplars as conditions to guide the model to super-resolve LR images. 
This is complemented by the proposed PWAve module, a component capable of learning how perform a suitable combination of the set of intermediate feature maps computed from the exemplars.
We empirically show the effectiveness of using more than one exemplar and that our method outperforms baselines from the literature. 
In addition, benefiting from the exemplars, our model can dynamically generate HR face images with arbitrary facial features.
\label{sec:Conclusion}

\bibliographystyle{splncs}
\bibliography{egbib}

\end{document}